\documentclass[letterpaper, 10 pt, conference]{ieeeconf}  

\IEEEoverridecommandlockouts                              

\let\labelindent\relax    
\usepackage[inline]{enumitem}  

\usepackage{amssymb}  

\usepackage{amsmath,amssymb}  
\usepackage{booktabs}         

\usepackage[hidelinks]{hyperref}
\usepackage[nameinlink,capitalize]{cleveref}

\usepackage[utf8]{inputenc}   
\usepackage{enumitem}

\usepackage{graphicx}   
\usepackage{subcaption} 

\usepackage[ruled,vlined,linesnumbered,noend]{algorithm2e}
\SetKwInput{KwInput}{Input}
\SetKwInput{KwOutput}{Output}
\SetKwProg{Fn}{Function}{}{}
\SetKw{KwAnd}{and}
\usepackage{amssymb}   
\usepackage{siunitx}   
\usepackage{wasysym}  

\title{PathFormer: A Transformer with 3D Grid Constraints for Digital Twin Robot-Arm Trajectory Generation}

\author{
Ahmed Alanazi$^{1}$, Duy Ho$^{2}$, and Yugyung Lee$^{1}$%
\thanks{$^{1}$Department of Computer Science, University of Missouri–Kansas City (UMKC), Kansas City, MO, USA.
{\tt\small \{aha85b, leeyu\}@umkc.edu}}%
\thanks{$^{2}$Department of Computer Science, California State University, Fullerton, CA, USA.
{\tt\small duyho@fullerton.edu}}%
\thanks{This work has been submitted to the IEEE International Conference on Robotics and Automation (ICRA) 2026 for review. Copyright may be transferred without notice, after which this version may no longer be accessible.}%
}

\begin{document}

\maketitle
\thispagestyle{empty}
\pagestyle{empty}

\begin{abstract}
Robotic arms require precise, task-aware trajectory planning, yet sequence models that ignore motion structure often yield invalid or inefficient executions. We present a \textbf{Path-based Transformer} that encodes robot motion with a \emph{3-grid (where/what/when) representation} and \emph{constraint-masked decoding}, enforcing lattice-adjacent moves and workspace bounds while reasoning over task graphs and action order. Trained on 53{,}755 trajectories (80\% train / 20\% validation), the model aligns closely with ground truth—\textbf{89.44\%} stepwise accuracy, \textbf{93.32\%} precision, \textbf{89.44\%} recall, and \textbf{90.40\%} F1—with \textbf{99.99\%} of paths legal by construction. Compiled to motor primitives on an \emph{xArm Lite\,6} with a depth-camera digital twin, it attains up to \textbf{97.5\%} reach and \textbf{92.5\%} pick success in controlled tests, and \textbf{86.7\%} end-to-end success across 60 language-specified tasks in cluttered scenes, absorbing slips and occlusions via local re-grounding without global re-planning. These results show that path-structured representations enable Transformers to generate accurate, reliable, and interpretable robot trajectories, bridging graph-based planning and sequence-based learning and providing a practical foundation for general-purpose manipulation and sim-to-real transfer.
\end{abstract}


\section{Introduction}
\label{sec:introduction}

Learning reliable policies for robot arms remains challenging as tasks grow more diverse and long-horizon. Recent progress in language-conditioned manipulation \cite{zhu2025equact,Gao2025MuST,mu2025look} shows that \emph{Transformers} can map high-level instructions to low-level actions, but most treat control as unconstrained sequence generation, ignoring geometric and kinematic structure. The result is brittle behavior: trajectories that work in-distribution can exhibit discontinuities, inefficiencies, or physically invalid steps when deployed.

These issues amplify in multi-step, multi-task settings, where a policy must not only produce the correct \emph{order} of actions but also ensure that each intermediate transition is \emph{feasible}—respecting lattice adjacency, workspace bounds, and subtask dependencies. Symbolic or finite-state scaffolds can enforce feasibility, but they rely on handcrafted abstractions that limit scalability. Pure sequence models excel at long-horizon reasoning, yet provide no guarantees that decoded trajectories remain valid—undermining their integration into \emph{digital twin} frameworks for safe planning, simulation, and sim-to-real transfer.

Recent digital twin efforts underscore both the opportunity and the gap. RoboTwin \cite{mu2025robotwin} highlights DTs for scalable data generation and evaluation; control-oriented studies pair reinforcement learning or hybrid controllers with DTs for real-time adjustment (e.g., SAC-based DT control for additive manufacturing \cite{mu2025robotwin} and PPO–fuzzy PID for arm stabilization \cite{cen2025digital}); perception-driven DTs maintain whole-body models for collision avoidance \cite{singh2024unity}. Across these, trajectory validity is typically enforced \emph{post hoc} via control corrections or environment updates, rather than guaranteed at generation time.

We introduce \textit{PathFormer}, a Transformer that generates robot-arm trajectories through a structured \emph{3-grid} representation. This design explicitly encodes (i) spatial transitions on a lattice (\emph{where}), (ii) task-graph dependencies among primitives (\emph{what}), and (iii) the sequential order of execution (\emph{when}). Decoding is \emph{graph-constrained}: at each step the model selects only from lattice-adjacent neighbors, guaranteeing legality by construction while retaining the expressivity of Transformer attention. The 3-grid aligns naturally with digital twins, enabling consistent simulation, safety checks, and transfer to physical arms without domain gaps.

\paragraph{Contributions.}
(1) \emph{3-grid trajectory representation:} a unified encoding of \emph{where/what/when} that supports joint reasoning over geometry, semantics, and order. 
(2) \emph{Graph-constrained Transformer decoding:} neighborhood masking restricts next-step predictions to valid lattice neighbors, providing trajectory-validity guarantees crucial for DT deployment.
(3) \emph{Empirical validation across diverse tasks:} on 53{,}755 trajectories, PathFormer attains \emph{89.44\%} stepwise accuracy, \emph{93.32\%} precision, and \emph{99.99\%} valid paths, and transfers to real robot arms with a depth-camera digital twin, executing \emph{reach}, \emph{pick}, and \emph{place} primitives.

By embedding structural constraints directly into Transformer-based generation, \emph{PathFormer} bridges sequence modeling and graph-based planning inside digital twins. The resulting policies are \emph{accurate}, \emph{interpretable}, and \emph{physically feasible}, supporting a wide spectrum of robot-arm tasks in unstructured environments.

\section{Related Work}
\label{sec:related}

\paragraph{Benchmarks for Robotic Manipulation}  
Large-scale simulators and benchmarks such as RLBench, Ravens, ALFRED, CALVIN, VLMbench, VIMA-Bench, and COLOSSEUM have accelerated progress in visuomotor learning by testing generalization across novel scenes, objects, and perceptual variations \cite{james2020rlbench,shridhar2020alfred,zeng2021transporter,mees2022calvin,zheng2022vlmbench,jiang2022vima,pumacay2024colosseum}.  
Recent additions like GemBench \cite{garcia2025towards} emphasize articulated objects and long-horizon task compositions.  
These platforms primarily measure robustness to perceptual or environmental variation, but they do not guarantee \emph{structural validity} of low-level trajectories.  
\textit{PathFormer} complements these efforts by shifting the focus from perceptual robustness to trajectory correctness, ensuring that every decoded path adheres to lattice and kinematic constraints.

\paragraph{Sequence Models for Manipulation}  
Transformer-based visuomotor policies and diffusion models achieve strong performance by directly generating long-horizon action sequences from demonstrations \cite{guhur2023instruction,shridhar2023perceiver,goyal2023rvt,chi2023diffusion,ke20253d}.  
However, since they optimize only for imitation fidelity, their outputs may contain discontinuities or physically invalid steps.  
\textit{PathFormer} closes this gap by embedding motion in a 3D lattice and applying constraint-masked decoding, thereby producing predictions that are both order-accurate and valid by construction.

\paragraph{LLM/VLM-Guided Planning}  
Language and vision--language models are increasingly used as high-level planners: SayCan \cite{brohan2023saycan} combines LLM reasoning with value functions, VoxPoser \cite{huang2023voxposer} encodes spatial affordances as value maps, and CaP \cite{liang2023code} generates robot-executable code.  
Recent work such as 3D-LOTUS++ \cite{garcia2025towards} integrates LLM reasoning with VLM-based grounding for open-vocabulary objects.  
These methods excel at symbolic task decomposition but still rely on low-level controllers that lack structural guarantees.  
\textit{PathFormer} provides a complementary execution layer: its lattice-constrained predictions ensure that trajectories invoked by high-level planners are always feasible and safe.

\paragraph{Structured Trajectory Representations and Digital Twins}  
Structured policies such as finite-state machines, task graphs, and option libraries offer interpretability but depend heavily on manual design \cite{jiang2022vima,pumacay2024colosseum}.  
Equivariant architectures \cite{zhu2025equact} and multi-skill Transformers \cite{Gao2025MuST} improve generalization but still decode trajectories without explicit feasibility checks.  
Meanwhile, digital twin approaches are increasingly used to provide safe, real-time mirroring of robot behavior \cite{mu2025robotwin,cen2025digital,singh2024unity}, supporting domains from dual-arm manipulation to manufacturing control.  
Similarly, knowledge-graph and GNN-based planners \cite{liu2024kg,ye2022comprehensive} capture structural dependencies and improve reasoning over multi-robot tasks.  
\textit{PathFormer} bridges these directions: by embedding trajectories in a 3D lattice and enforcing graph-constrained decoding, it combines the rigor of graph-based planning with the scalability of modern sequence models, and can be integrated naturally with digital twins or KG/GNN-based frameworks.

\paragraph{Summary}  
Benchmarks advance evaluation, Transformers advance multi-skill visuomotor learning, and LLM/VLM systems advance task-level reasoning.  
Yet ensuring that generated \emph{trajectories} remain valid under geometric and kinematic constraints is still unresolved.  
\textit{PathFormer} addresses this gap: it produces robot motions that are \emph{valid by construction}, interpretable via a 3D grid, and transferable to both digital twins and physical robot arms, while remaining compatible with higher-level KG- or LLM-driven planners.

\section{Methodology}
\label{sec:method}

\begin{figure}[t]
  \centering
  \includegraphics[width=\linewidth]{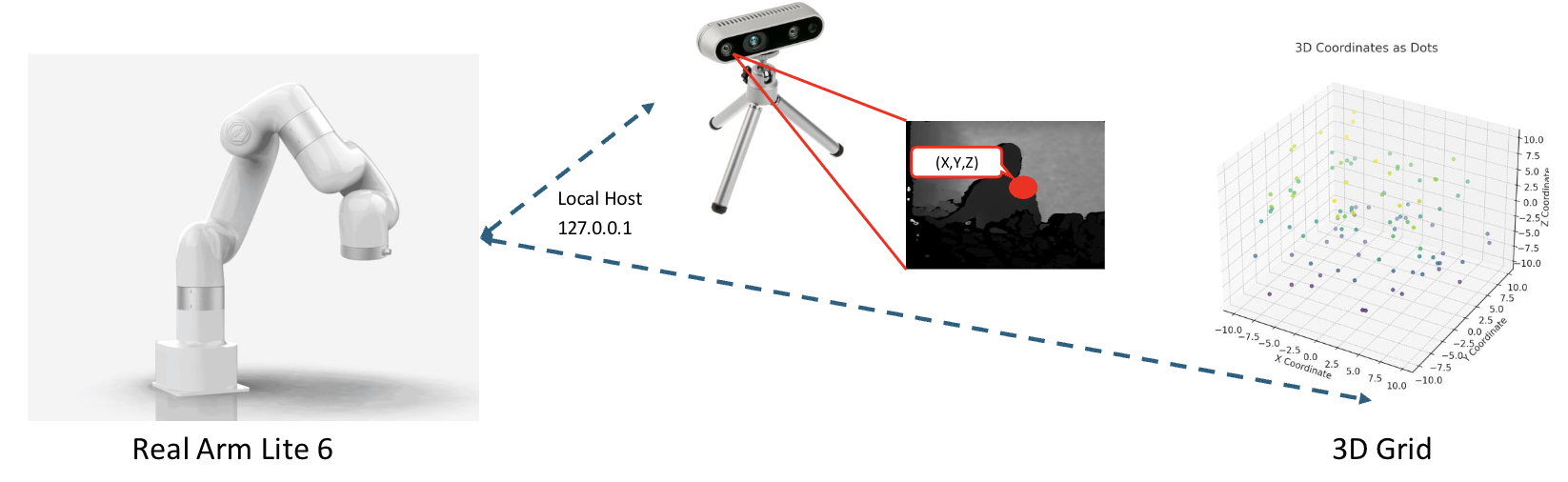}
  \caption{\textbf{Digital-twin sensing to grid.}
  A RealSense depth camera provides $(x,y,z)$ targets on the local host; detections are projected onto a 3D grid used by PathFormer.
  The Lite~6 arm executes plans generated on this grid, enabling consistent sim-to-real transfer.}
  \label{fig:dt-setup}
\end{figure}

\begin{figure}[t]
  \centering
  \includegraphics[width=\linewidth]{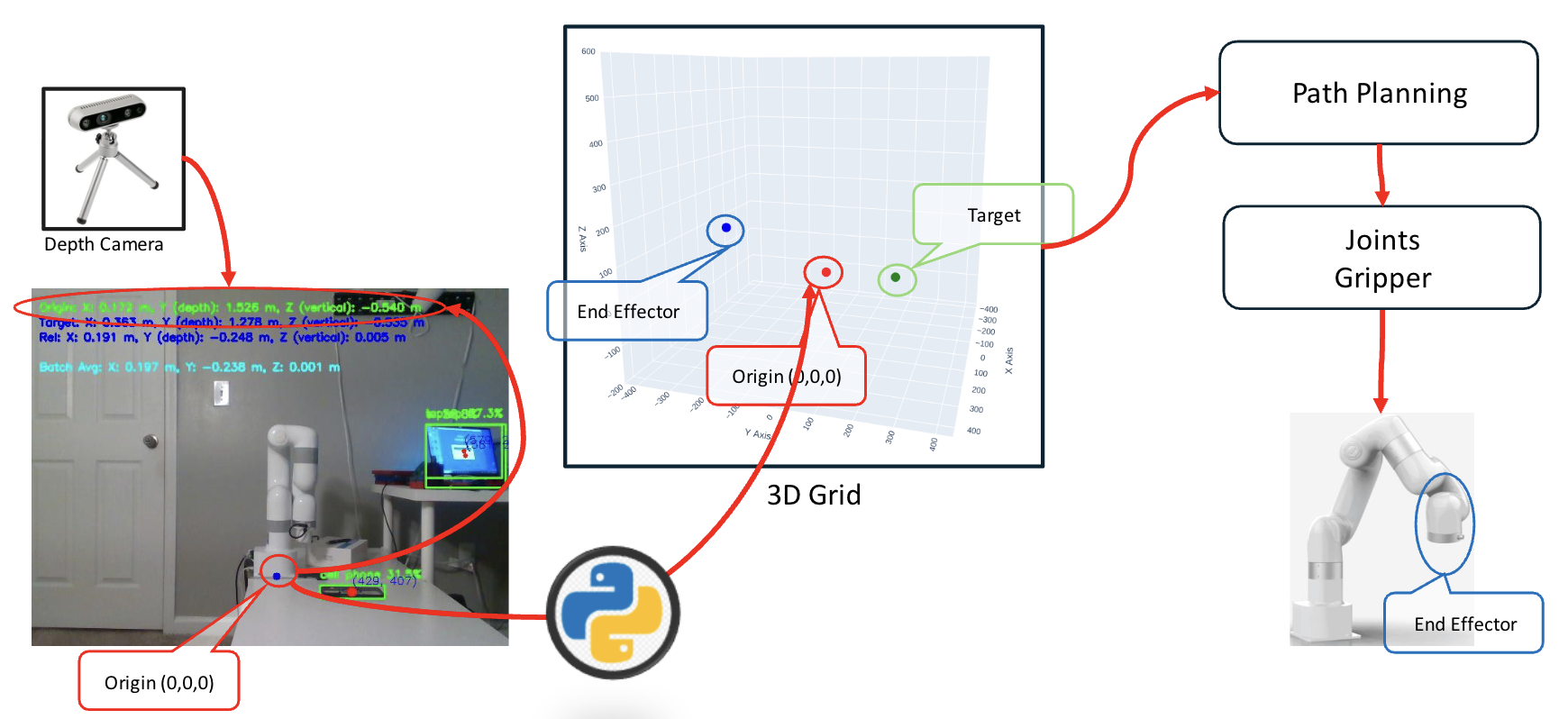}
  \caption{\textbf{End-to-end pipeline.}
  Depth camera $\rightarrow$ 3D grid (origin, target, end-effector) $\rightarrow$ \emph{PathFormer} graph-constrained decoding
  $\rightarrow$ joint/gripper commands. The grid synchronizes the digital twin with the physical arm, while neighborhood masking enforces lattice adjacency.}
  \label{fig:dt-pipeline}
\end{figure}

\begin{figure*}[t]
  \centering
  \includegraphics[width=\textwidth]{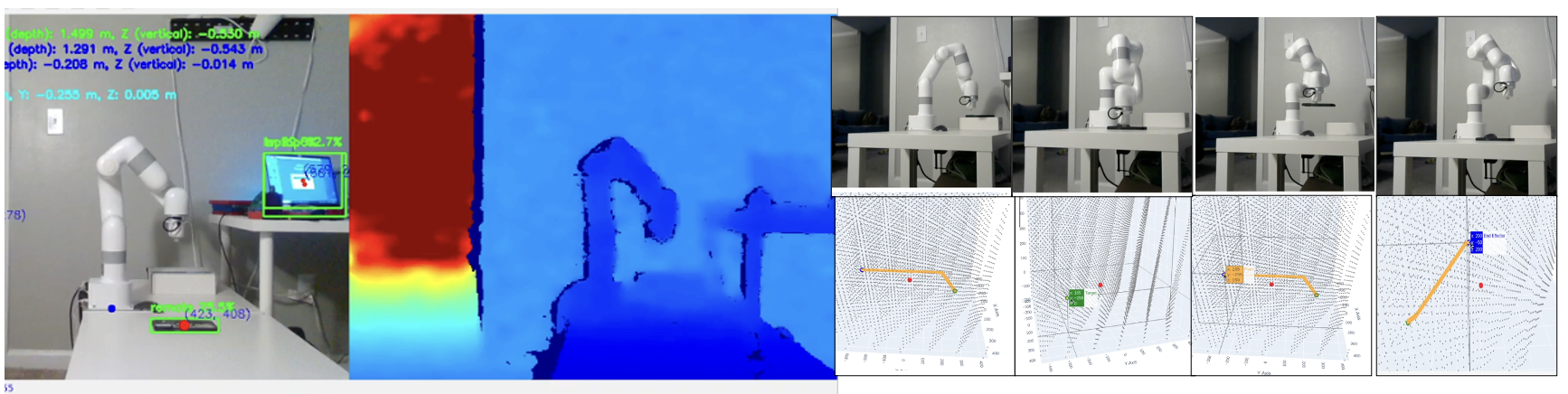}
  \caption{\textbf{Perception-to-execution rollout.}
  \emph{Left:} RGB and depth streams from the RealSense D435i with YOLO-based object detection. 
  \emph{Middle:} Lite~6 robot executing \textit{reach--pick--place}. 
  \emph{Right:} decoded 3D-grid trajectories for corresponding task phases. 
  Together, these demonstrate how PathFormer integrates perception, digital twin alignment, and real-robot execution.}
  \label{fig:percep-exec}
\end{figure*}

\begin{figure*}[t]
  \centering
  \includegraphics[width=\textwidth]{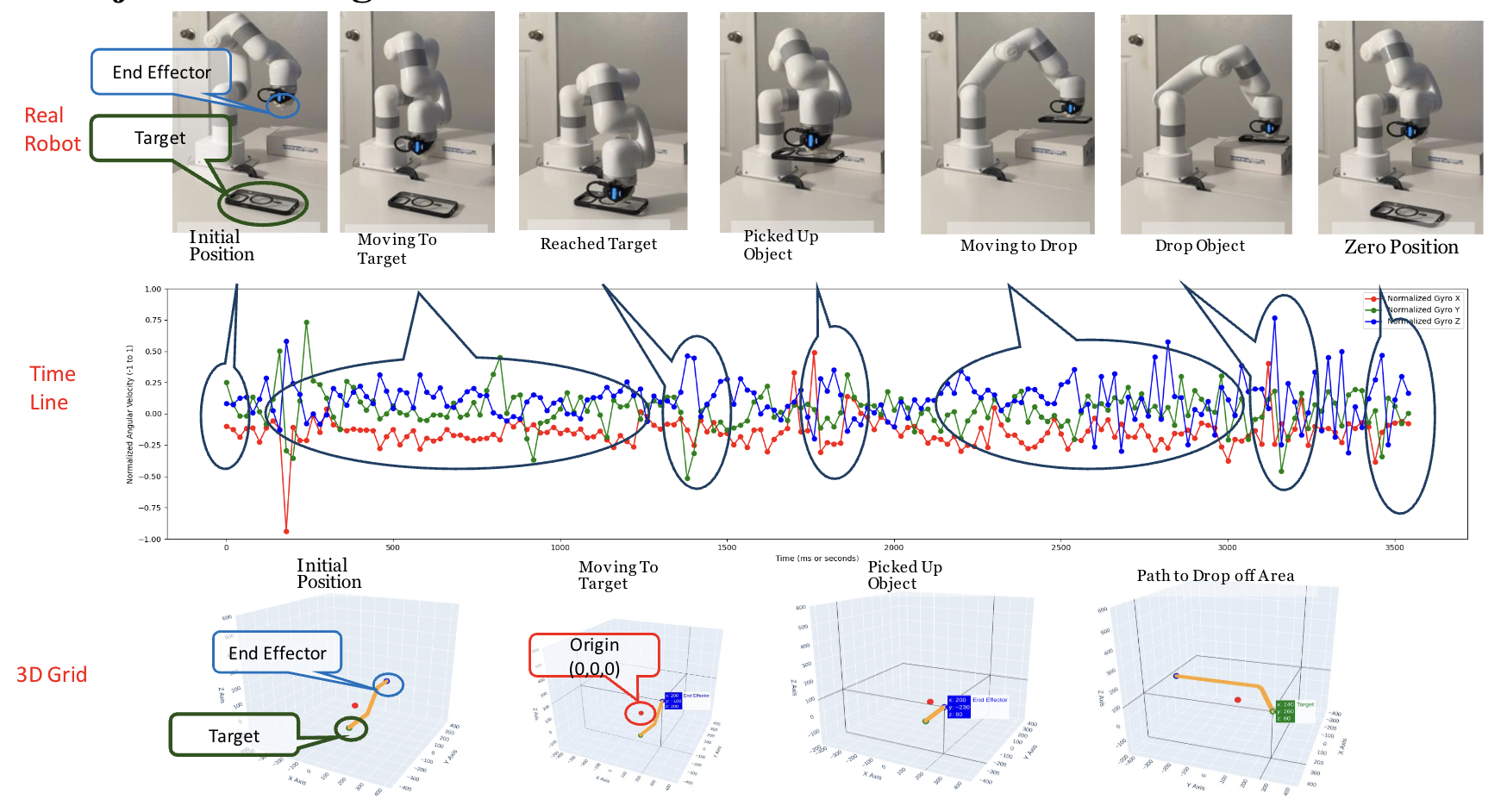}
  \caption{\textbf{Real-robot rollout and grid trajectories.}
  \emph{Top:} snapshots of \textit{reach--pick--place}, showing end-effector motion across phases.  
  \emph{Middle:} normalized wrist IMU/gyro timeline, with annotations marking transitions between subtasks.  
  \emph{Bottom:} decoded 3D-grid paths (origin, target, end-effector) aligned with each phase.  
  These results confirm that the digital twin mirrors physical execution while lattice adjacency constraints prevent discontinuities.}
  \label{fig:arm-time-grid}
\end{figure*}

We propose \textit{PathFormer}, a \textit{causal Transformer} that generates robot-arm trajectories on a discrete lattice using a structured \emph{3D grid} representation. Unlike prior sequence models that decode unconstrained actions \cite{guhur2023instruction,shridhar2023perceiver,chi2023diffusion}, PathFormer couples a depth-camera \emph{digital twin} with \textit{graph-constrained decoding}, ensuring trajectories are both \emph{accurate} and \emph{valid by construction}. The sensing-to-grid setup, the end-to-end pipeline, and perception-to-execution rollout are shown in Figs.~\ref{fig:dt-setup}--\ref{fig:percep-exec}. A detailed real-robot rollout is presented later (Fig.~\ref{fig:arm-time-grid}).

\paragraph{Relation to Transformer planners.}
Transformer-based task planners often embed graph structure at the \emph{task level}. For instance, the GNN-Transformer Task Planner (GTTP) integrates scene-graph encoders with Transformer layers to predict subgoals from language instructions \cite{jeong2025gnn}. PathFormer differs fundamentally: it operates at the \emph{trajectory level}, autoregressively decoding \emph{lattice coordinates} with a neighborhood mask that enforces Manhattan adjacency and workspace bounds. Thus, PathFormer complements GTTP-style planners—task-level planners can propose subgoals or skills, while PathFormer guarantees \emph{feasible low-level execution} of those subgoals on real hardware.

\subsection{Problem Formulation}
A trajectory is a path $\pi=\{p_1,\dots,p_T\}$ on a 3D lattice:
\begin{equation}
p_t=(x_t,y_t,z_t),\quad p_t\in\mathbb{Z}^3,\quad
\|p_{t+1}-p_t\|_1 = 1,
\end{equation}
where each move is lattice-adjacent. Given an initial state $p_0$ from the robot’s end effector and a task context $C$ (task-graph features, Sec.~\ref{sec:three_grid}), PathFormer predicts $\pi$ that (i) matches ground-truth order (\emph{Stepwise Accuracy}), (ii) recovers the correct set of coordinates (\emph{Precision/Recall/F1}), and (iii) respects adjacency and workspace bounds (\emph{Validity}).

\subsection{3-Grid Representation and Spatial Neighborhood}
\label{sec:three_grid}
We encode trajectories with three complementary grids:
\begin{itemize}
\item \textit{Spatial Grid} $\mathcal{G}_{\text{spatial}}=(\mathcal{V},\mathcal{E})$: nodes $\mathcal{V}$ index lattice coordinates; edges connect neighbors
\begin{equation}
(v_i,v_j)\in\mathcal{E}\iff \|p_i-p_j\|_1=1 \ \text{and}\ p_j\in\Omega,
\end{equation}
where $\Omega$ denotes workspace bounds (derived from the Lite6 robot’s 440mm reach and obstacle masks from the depth camera). This defines the legal action set at each step.
\item \textit{Task Grid}: a DAG over sub-tasks (e.g., grasp, lift, place) with dependency edges (pre$\rightarrow$post), mapped to tokens/features $C$.
\item \textit{Sequence Grid}: encodes temporal indices and causal masks to align positions with action order.
\end{itemize}

\noindent\textit{Knowledge-Graph View.} The combined structure can be seen as a knowledge graph $\mathcal{G}=(\mathcal{V},\mathcal{E},\mathcal{A})$ whose attributes $\mathcal{A}$ annotate nodes/edges with action labels, step indices, and task tags. A trajectory is thus a path traversal on $\mathcal{G}$.

\subsection{Digital Twin Pipeline}
To synchronize real-world sensing with simulation (Fig.~\ref{fig:dt-pipeline}):
\begin{enumerate}
\item A depth camera (Intel RealSense) localizes objects in $(x,y,z)$ using YoloV11.
\item Coordinates are projected into a voxelized 3D grid, where the origin $(0,0,0)$ aligns with the robot base, the target is inserted, and the end effector is tracked.
\item The Lite6 robot arm (6 DoF, 440mm reach, $\pm0.5$mm repeatability) executes candidate paths, while the digital twin validates feasibility in parallel.
\item Feedback from both environments keeps the spatial grid consistent and provides training/evaluation data for PathFormer.
\end{enumerate}

\subsection{Inputs and Embeddings}
For each candidate cell $u\in\mathcal{V}$, we build an embedding:
\begin{equation}
e(u)=E_{\text{coord}}(x,y,z)\;+\;E_{\text{task}}(C[u])\;+\;E_{\text{seq}}(t),
\end{equation}
where $E_{\text{coord}}$ encodes spatial coordinates, $E_{\text{task}}$ encodes sub-task features (e.g., grasp vs.~place), and $E_{\text{seq}}$ encodes step index $t$. This fuses geometry, semantics, and temporal order into a shared latent space.

\subsection{Transformer with Constraint Masking}
\label{sec:arch}
We use a causal Transformer decoder to predict $p_{t+1}$ given $\{p_1,\dots,p_t\}$ and context $C$. A \emph{neighborhood mask} $M_t$ restricts logits to valid neighbors:
\begin{equation}
M_t(u)=\mathbb{1}[u\in\mathcal{N}(p_t)], \quad 
\mathcal{N}(p_t)=\{u \mid (p_t,u)\in\mathcal{E}\}.
\end{equation}
\noindent Validated logits are:
\begin{equation}
\tilde{\ell}(u)=
\begin{cases}
\ell(u), & M_t(u)=1\\
-\infty, & M_t(u)=0
\end{cases}, \quad
p_{t+1}=\arg\max_u \text{softmax}(\tilde{\ell}(u)).
\end{equation}
This guarantees lattice-adjacent motion and bounded feasibility by construction.

\subsection{Training Objective}
We train on a 53k-scale corpus of trajectories  with a composite loss:
\begin{equation}
\mathcal{L}=\mathcal{L}_{\text{seq}}
+\lambda_1 \mathcal{L}_{\text{coord}}
+\lambda_2 \mathcal{L}_{\text{valid}}
+\lambda_3 \mathcal{L}_{\text{cov}}
+\lambda_4 \mathcal{L}_{\text{len}},
\end{equation}
where $\mathcal{L}_{\text{seq}}$ supervises next-step indices, $\mathcal{L}_{\text{coord}}$ optimizes coordinate set recall/precision, $\mathcal{L}_{\text{valid}}$ penalizes invalid moves, $\mathcal{L}_{\text{cov}}$ encourages full path coverage, and $\mathcal{L}_{\text{len}}$ regularizes path length.

\subsection{Constrained Decoding (Pseudocode)}
\begin{algorithm}[htbp]
\caption{Constrained decoding on $\mathcal{G}_{\text{spatial}}$}
\label{alg:decode}
\KwIn{start $p_0$, context $C$, max steps $T_{\max}$}
$\pi\leftarrow [\ ]$; $p\leftarrow p_0$\;
\For{$t=0$ \KwTo $T_{\max}-1$}{
  $h_t\leftarrow \textsc{Transformer}( \pi, C )$\;
  $\ell\leftarrow W h_t + b$ \tcp*{logits}
  $\mathcal{N}\leftarrow \{u\mid \|u-p\|_1=1,\ u\in\Omega\}$\;
  $\tilde{\ell}(u)\leftarrow \ell(u)$ if $u\in\mathcal{N}$ else $-\infty$\;
  $p'\leftarrow \arg\max_u \text{softmax}(\tilde{\ell}(u))$\;
  \If{$p'$ is terminal or $p'=p$}{\textbf{break}}
  $\pi.\text{append}(p')$; $p\leftarrow p'$
}
\Return $\pi$
\end{algorithm}

\subsection{Implementation Platform}
\label{sec:platform}

\paragraph{Hardware}  
Experiments are conducted on \textit{UFactory collaborative manipulators} (\emph{xArm5/6/7}, Lite6, and UFACTORY850), supporting interchangeable end-effectors (parallel-jaw, vacuum, and BIO grippers). Dual-arm configurations run in a single \texttt{rviz} instance with independent DoFs and synchronized task graphs.

\begin{figure}[t]
    \centering
    \begin{subfigure}[b]{0.45\linewidth}
        \centering
        \includegraphics[width=\linewidth]{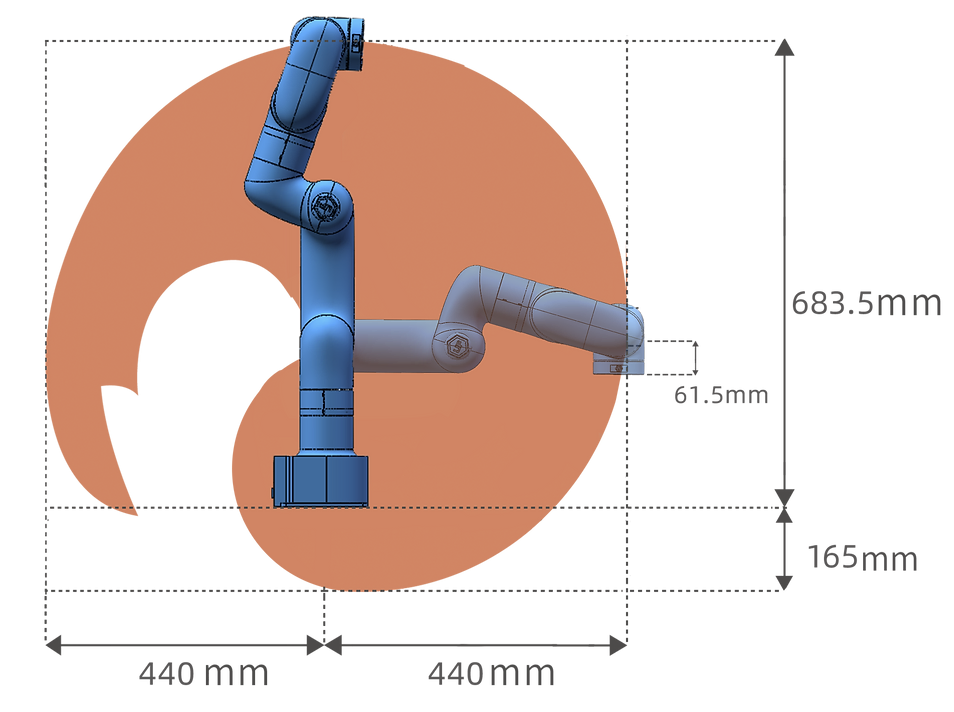}
        \caption{Robot Arm dimensions (side view).}
        \label{fig:arm-dim-side}
    \end{subfigure}
    \hfill
    \begin{subfigure}[b]{0.45\linewidth}
        \centering
        \includegraphics[width=\linewidth]{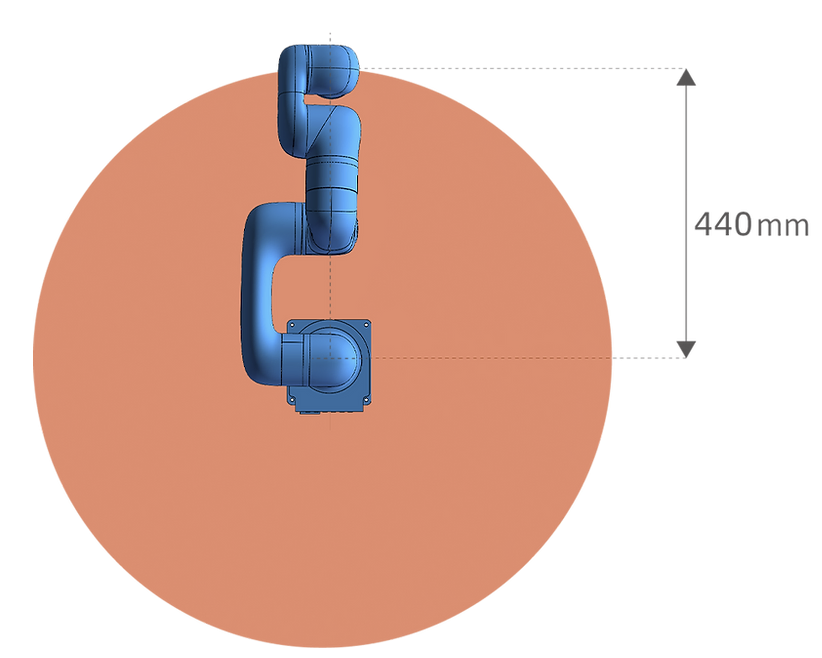}
        \caption{Robot Arm workspace radius (top view).}
        \label{fig:arm-dim-top}
    \end{subfigure}

    \begin{subfigure}[b]{0.55\linewidth}
        \centering
        \includegraphics[width=\linewidth]{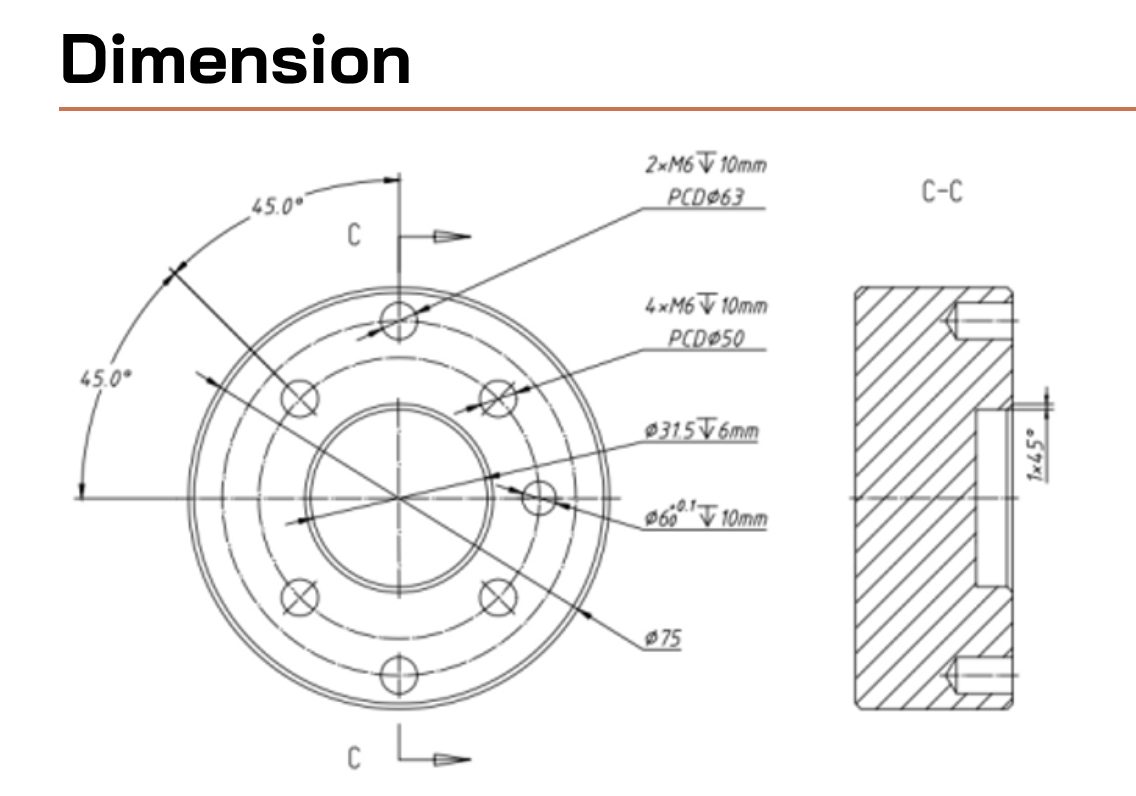}
        \caption{End-effector mount dimension diagram.}
        \label{fig:end-effector-dim}
    \end{subfigure}
    \hfill
    \begin{subfigure}[b]{0.40\linewidth}
        \centering
        \includegraphics[width=\linewidth]{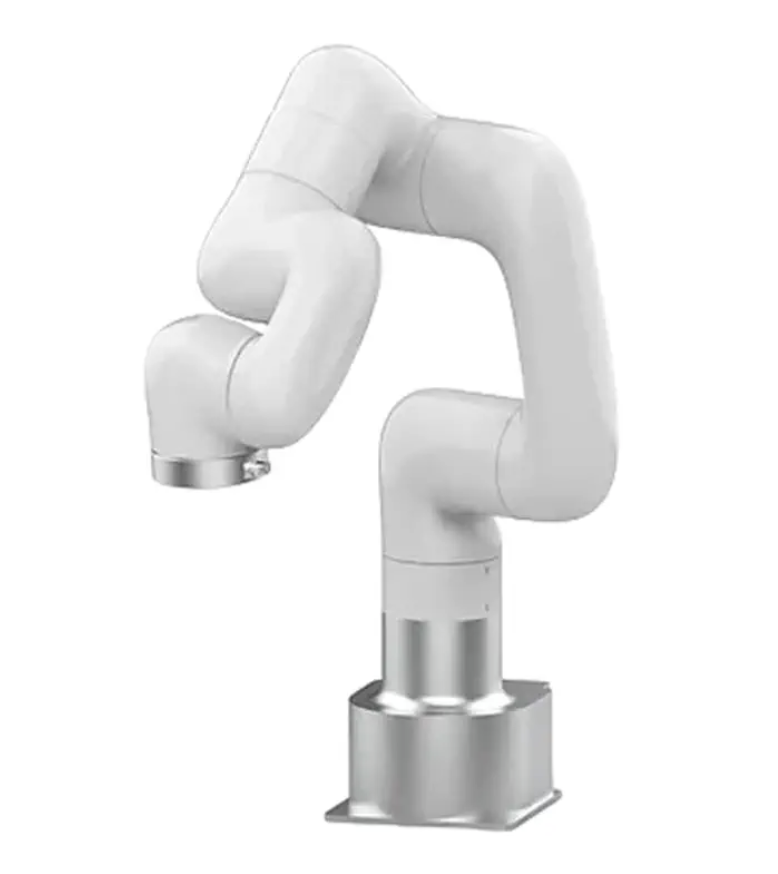}
        \caption{Physical robot arm (6-DoF).}
        \label{fig:robot-arm-photo}
    \end{subfigure}

    \caption{Robot Arm hardware specifications and dimensions.}
    \label{fig:robot-arm}
\end{figure}

\paragraph{Operating Systems and Middleware}  
All experiments run on \textit{Ubuntu LTS} with \textit{ROS~2} releases (\emph{Foxy}, \emph{Galactic}, \emph{Humble}, \emph{Jazzy}, and \emph{Rolling}), ensuring vendor-branch API compatibility.  
The control stack builds on the BSD-licensed \texttt{xarm\_ros2} meta-repo, summarized below.

\begin{table}[ht]
\centering
\caption{ROS~2 middleware in the \texttt{xarm\_ros2} meta-repo.}
\label{tab:ros2_middleware}
\begin{tabular}{p{5cm}p{3.7cm}}
\toprule
\textbf{Package} & \textbf{Function} \\
\midrule
\texttt{xarm\_description} & URDF/XACRO robot models \\
\texttt{xarm\_api}, \texttt{xarm\_sdk} & HW services and ROS topics \\
\texttt{xarm\_controller} & Hardware interface layer \\
\texttt{xarm\_moveit\_config}, \texttt{xarm\_planner} & MoveIt~2 motion planning \\
\texttt{xarm\_gazebo} & Gazebo simulation \\
\texttt{xarm\_moveit\_servo} & Teleoperation / servo control \\
\texttt{xarm\_vision} & RealSense drivers and calibration \\
\bottomrule
\end{tabular}
\end{table}

\subsection{Robot Geometry and Workspace}
\label{sec:workspace}

\paragraph{Reach Envelope}  
The Lite6 workspace, derived from CAD views (Figs.~\ref{fig:arm-dim-side}, \ref{fig:arm-dim-top}), is summarized in Table~\ref{tab:reach_envelope}.  

\begin{table}[ht]
\centering
\caption{Lite6 reach envelope dimensions.}
\label{tab:reach_envelope}
\begin{tabular}{p{3.2cm}p{4.5cm}}
\toprule
\textbf{Parameter} & \textbf{Value} \\
\midrule
Planar radius & \textit{440\,mm} (span $\approx$880\,mm) \\
Vertical envelope & \textit{683.5\,mm} (base to highest reach) \\
Offsets & Base: \textit{165\,mm}, TCP: \textit{61.5\,mm} \\
\bottomrule
\end{tabular}
\end{table}

These dimensions define the lattice bounds for PathFormer:
\[
x,y \in [-440, 440]~\text{mm}, \qquad z \in [0, 683.5]~\text{mm}.
\]

\paragraph{Mounting Interface}  
The flange geometry (Fig.~\ref{fig:end-effector-dim}) is summarized in Table~\ref{tab:mounting_interface}. These dimensions ensure precise registration of end-effectors and alignment of the lattice frame with the TCP.

\begin{table}[ht]
\centering
\caption{Lite6 end-effector flange specifications.}
\label{tab:mounting_interface}
\begin{tabular}{p{3.2cm}p{4.5cm}}
\toprule
\textbf{Parameter} & \textbf{Value} \\
\midrule
Mounting face & \textit{\diameter 75\,mm} \\
Bolt circles & $2\times$ M6 @ PCD~63\,mm; $4\times$ M6 @ PCD~50\,mm \\
Bore / Counterbore & \diameter 31.5\,mm / \diameter 60\,mm, with \ang{45} chamfers \\
\bottomrule
\end{tabular}
\end{table}

\paragraph{Lattice Parameterization}  
The workspace is discretized at a resolution of \textit{20\,mm} per axis, yielding:
\[
\mathcal{V}=\Bigl\{(x,y,z)\in \mathbb{Z}^3 \;\big|\;
   x,y \in [-22,\dots,22],\; z\in[0,\dots,34] \Bigr\}.
\]

Constraint-masked decoding enforces lattice adjacency:
\[
\ell_1(p_t, p_{t+1}) = 1, \qquad p_{t+1} \in \mathcal{V},
\]
where $\ell_1$ is the Manhattan distance. The TCP offset (\textit{61.5\,mm}) is applied as a fixed transform.

\subsection{Perception, Calibration, and Safety}
A \textit{RealSense D435i} provides depth sensing, with automated hand–eye calibration (\texttt{aruco}/\texttt{easy\_handeye2}). Extrinsics are persisted and reloaded at launch. \textit{MoveIt~2} enforces joint limits, self-collision, and planning-scene checks. YOLOv11-based detection supplies $(x,y,z)$ targets for digital twin updates.

\subsection{PathFormer Integration with Digital Twins}
PathFormer executes as a ROS~2 node, consuming task context and state (from Gazebo/MoveIt or \texttt{joint\_states}) and producing lattice-valid waypoints:
\begin{enumerate}[leftmargin=10pt,itemsep=2pt,topsep=1pt]
  \item Validity-checked against workspace bounds,
  \item Time-parameterized by MoveIt~2 for smooth execution, or sent directly via \texttt{xarm\_api}.
\end{enumerate}

In digital twin mode (Fig.~\ref{fig:dt-setup}), the depth-camera stream synchronizes with the 3D grid in real time, supporting both single-arm and dual-arm experiments.

\section{Experimental Results}
\label{sec:results}

We evaluate \textit{PathFormer} with \textit{three experimental designs} spanning simulation and hardware:

\noindent\textit{D1 — Lattice-trajectory decoding (offline).} Evaluation on \textit{10{,}751} labeled lattice trajectories to test \emph{order correctness} and \emph{coordinate-set recovery} (Table~\ref{tab:main_results}).  

\noindent\textit{D2 — Robotic trajectory execution (controlled bench).} \emph{xArm Lite\,6} paired with a depth-camera digital twin; lattice paths mapped to grasp--move--release primitives to assess feasibility, stability, and mechanical failure modes under controlled layouts.  

\noindent\textit{D3 — Real-world end-to-end manipulation (cluttered scenes).} Natural-language pickup--place tasks on the same platform with perception, constraint checks, and a transformer policy; measures sim$\to$real transfer and robustness in dynamic, occluded settings (Tables~\ref{tab:robot_results}, \ref{tab:failures}).  

These studies address:  
(Q1) Are trajectories \emph{order-correct}?  
(Q2) Do they recover the \emph{right set} of coordinates?  
(Q3) Are predictions \emph{valid by construction} and transferable to physical hardware?

\subsection{Evaluation Metrics}
We evaluate PathFormer using four complementary metrics:

\paragraph{Stepwise Accuracy.}  
For a predicted trajectory $\hat{\pi} = \{\hat{p}_1, \ldots, \hat{p}_T\}$ and ground-truth $\pi^\star = \{p^\star_1, \ldots, p^\star_T\}$:
\[
\mathrm{Acc}_{\text{step}} = \frac{1}{T}\sum_{t=1}^{T} \mathbb{1}\left[\hat{p}_t = p^\star_t\right].
\]
This order-sensitive metric penalizes truncations, swaps, or over-extensions. PathFormer achieves \textit{89.44\%}, i.e., $\sim$9/10 steps are aligned with the gold trajectory.

\paragraph{Coordinate Precision, Recall, and F1.}  
Order-agnostic set overlap between predictions and ground truth is measured by:
\[
\mathrm{Precision} = \tfrac{|\hat{\Pi}\cap \Pi^\star|}{|\hat{\Pi}|},\quad 
\mathrm{Recall} = \tfrac{|\hat{\Pi}\cap \Pi^\star|}{|\Pi^\star|},\quad
\mathrm{F1} = \tfrac{2PR}{P+R}.
\]
PathFormer achieves \textit{93.32\%} precision, \textit{89.44\%} recall, and \textit{90.40\%} F1—demonstrating strong balance between correctness and coverage.

\paragraph{Valid Path Percent.}  
We compute the fraction of predictions that obey lattice adjacency and workspace bounds:
\[
\mathrm{Valid} = \frac{1}{N}\sum_{i=1}^{N}\mathbb{1}\big[\forall t,\;\|\hat{p}_{t+1}-\hat{p}_{t}\|_1=1 \wedge \hat{p}_t \in \mathcal{V}\big].
\]
PathFormer achieves \textit{99.99\%}, indicating that virtually all decoded paths are legal by construction.

\paragraph{Real-Robot Success Rate.}  
Physical execution success is defined as completing a primitive without collision or boundary violation:
\[
\mathrm{Success} = \frac{\# \; \text{successful trials}}{\# \; \text{total trials}} \times 100\%.
\]
Results peak at \textit{97.5\%} for reaching and \textit{92.5\%} for picking in central workspace ranges, with mild degradation at the periphery.

\subsection{Results by Experimental Design}
\label{subsec:results_by_design}

\paragraph{Setup (summary).}
All experiments use the same decoding stack and safety envelopes across simulation and hardware. The \emph{lattice decoder} emits unit-adjacent moves with constraint masks; a \emph{digital twin} maintains a time-synchronized scene graph (pose grids, occlusion map, reachability) and streams targets to an \emph{xArm Lite\,6} controller with joint-velocity commands, geofences, depth/angle gates, and max-force abort. Depth images are registered to the robot base with a one-time extrinsic calibration; predicted trajectories are compared to measured end-effector traces after tool-center-point (TCP) offset compensation.

\paragraph{D1 — Lattice-trajectory decoding (offline).}
We quantify \emph{order correctness}, \emph{coordinate-set recovery}, and \emph{legality} on $10{,}751$ labeled trajectories (no scene images; pure symbolic decoding).
Decoding uses beam width $B{=}5$ with legality masks; ties are broken by coverage penalties to discourage early truncation.
\vspace{2pt}

\begin{table}[ht]
\centering
\caption{Lattice corpus results (trajectory fidelity and validity).}
\label{tab:main_results}
\begin{tabular}{lcccc}
\toprule
\textbf{Metric} & \textbf{Stepwise Acc.} & \textbf{Precision} & \textbf{Recall} & \textbf{F1} \\
\midrule
Value (\%) & 89.44 & 93.32 & 89.44 & 90.40 \\
\midrule
\multicolumn{5}{c}{\textit{Validity}} \\
\midrule
Valid Path Percent & \multicolumn{4}{c}{99.99} \\
\bottomrule
\end{tabular}
\end{table}

\noindent\emph{Error profile.}
Most residual errors are (E1) tail truncation on long horizons and (E2) adjacent-step swaps;
(E3) boundary nudges occur when alternatives with equal score choose different but still valid neighbors.
Legality masks eliminate illegal jumps (L1) by construction.

\paragraph{D2 — Robotic trajectory execution (controlled bench).}
Lattice paths are compiled to \emph{grasp--move--release} primitives with phase-conditioned gains.
The controller executes ``Approach $\to$ Engage $\to$ Transport $\to$ Release'' with impedance/admittance tracking;
contact spikes trigger slow-down or abort, and geofences prevent out-of-workspace motion.
In this setting, \emph{graph-constrained decoding} reliably grounds symbolic plans into feasible motor programs.
Failures are rare and predominantly mechanical (e.g., transient gripper slip on glossy plastic).
When minor object shifts are detected (e.g., a $15^\circ$ cube rotation), the twin re-grounding updates targets and execution proceeds without global re-planning.

\begin{figure}[ht]
    \centering
    \includegraphics[width=0.95\linewidth]{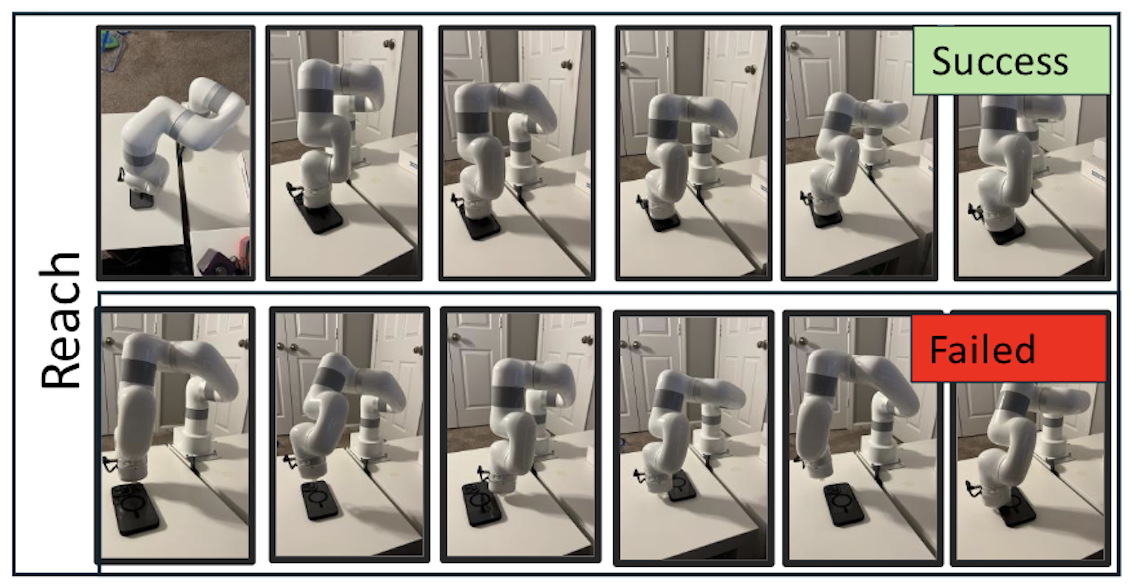}
    \caption{D2: Real-robot \emph{Reach} task—success (top) and failure (bottom).}
    \label{fig:reach}
\end{figure}

\begin{figure}[ht]
    \centering
    \includegraphics[width=0.95\linewidth]{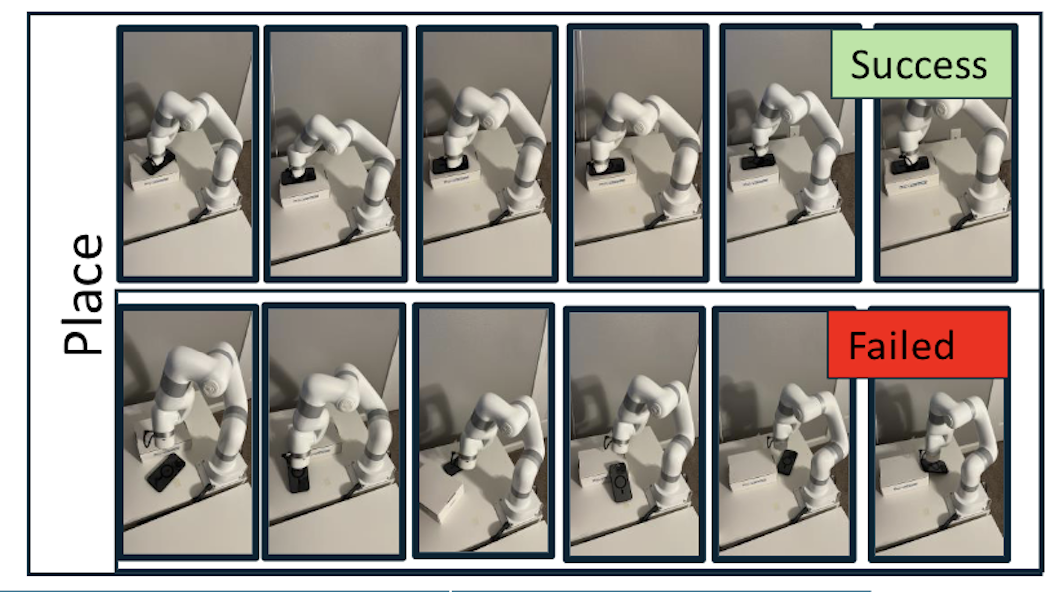}
    \caption{D2: Real-robot \emph{Place} task—success (top) and failure (bottom).}
    \label{fig:place}
\end{figure}

\begin{table}[ht]
\centering
\caption{D2: Real-robot success rates (SR, \%) for \emph{Reach} and \emph{Pick} across workspace ranges.}
\label{tab:rr_success}
\begin{tabular}{p{1.1cm}p{5cm}c}
\toprule
\textbf{Task} & \textbf{Area $(x,y,z)$ [mm]} & \textbf{SR} \\
\midrule
Reach & Range 1: $(190$--$210,\,-225$--$235,\,0$--$10)$ & 97.5 \\
      & Range 2: $(220$--$235,\,-170$--$185,\,0$--$10)$ & 95.0 \\
      & Range 3: $(240$--$255,\,-200$--$215,\,0$--$10)$ & 95.0 \\
\midrule
Pick  & Range 1: $(190$--$210,\,-225$--$235,\,0$--$10)$ & 92.5 \\
      & Range 2: $(220$--$235,\,-170$--$185,\,0$--$10)$ & 87.5 \\
      & Range 3: $(240$--$255,\,-200$--$215,\,0$--$10)$ & 90.0 \\
\bottomrule
\end{tabular}
\end{table}

\noindent\emph{Interpretation.}
Central workspace (Range~1) yields the highest success; slight degradation near Ranges~2/3 aligns with kinematic reach and camera perspective foreshortening.
In successful runs, predicted paths overlay measured end-effector trajectories; residuals reflect boundary nudges (camera quantization, TCP offsets).

\paragraph{D3 — Real-world end-to-end manipulation (cluttered scenes).}
We evaluate \textit{60} natural-language pickup--place instructions (e.g., \texttt{pickup smartphone $>$ place in container}) in dynamic, partially occluded scenes using (i) LLM-based perception, (ii) feasibility/occlusion checks, and (iii) a transformer policy conditioned on scene state and constraints.

\noindent\emph{Unexpected events \& local re-grounding.}
\begin{itemize}
  \item \textit{Slipped object:} a phone slid $\sim$4\,cm after a near-miss; pose was re-grounded and the next grasp succeeded without discarding the plan.
  \item \textit{Occlusion:} a partially hidden cup caused an invalid first grasp; visibility/approach updated and the second attempt completed.
  \item \textit{Dynamic obstacle:} a spoon drifted into the reach path; a one-cell detour preserved lattice feasibility and rejoined the nominal route.
\end{itemize}

\begin{table*}[t]
\centering
\caption{D3: Real-world results on 60 instructions (cluttered tabletop).}
\label{tab:robot_results}
\begin{tabular}{lcc}
\toprule
\textbf{Metric} & \textbf{Value (\%)} & \textbf{Notes} \\
\midrule
Successful Executions & \textbf{86.7} & 52/60 completed \\
Grasp Success Rate    & \textbf{90.0} & Most failures from occlusion/tight packing \\
Placement Accuracy    & \textbf{93.3} & Minor drop-location deviations \\
\bottomrule
\end{tabular}
\end{table*}

\noindent\textit{Failure analysis (8/60; 13.3\%).}
Four dominant modes were observed; counts are summarized in Table~\ref{tab:failures}.
\begin{itemize}
  \item \textit{State-sync ``no state''} (heavy occlusion or ambiguous ``an object'' prompts). \textit{Mitigation:} target disambiguation + retry perception sweep.
  \item \textit{Occlusion/tight clustering} (blocked approach inside the container). \textit{Mitigation:} local clearing or alternate grasp.
  \item \textit{Small nested item blocks grasp}. \textit{Mitigation:} nested-occupancy pre-check; reorder picks.
  \item \textit{Perception mis-ID} (category/instance error). \textit{Mitigation:} frame-majority voting; confidence-gated execution.
\end{itemize}

\begin{table}[ht]
\centering
\caption{D3: Failure modes and counts (60 trials).}
\label{tab:failures}
\begin{tabular}{lcc}
\toprule
\textbf{Failure mode} & \textbf{Count} & \textbf{Rate (\%)} \\
\midrule
No state available (state-sync) & 5 & 8.3 \\
Occlusion / tight clustering    & 1 & 1.7 \\
Small object blocks grasp       & 1 & 1.7 \\
Perception mis-prediction       & 1 & 1.7 \\
\midrule
\textbf{Total}                  & \textbf{8} & \textbf{13.3} \\
\bottomrule
\end{tabular}
\end{table}

\paragraph{Takeaways.}
(1) \emph{Symbolic$\to$motor reliability:} lattice plans ground into phase-conditioned motor primitives across layouts. 
(2) \emph{Robustness in clutter:} end-to-end success $\sim 87\%$ with local re-grounding \emph{without} global re-plans. 
(3) \emph{Adaptivity:} slips, occlusions, and incidental obstacles are handled via pose re-estimation and one-cell detours. 
(4) \emph{Roadmap:} failures motivate multi-view sensing, target disambiguation, grasp reordering, and confidence-gated execution.

\subsection{Discussion}

\noindent\textbf{Summary across D1--D3.}
The Path-based Transformer demonstrates that trajectory generation can be simultaneously \emph{data-driven} and \emph{structurally guaranteed} across three settings: 
\emph{D1} (offline lattice decoding), \emph{D2} (controlled robotic execution with a digital twin), and \emph{D3} (real-world, cluttered manipulation).
On the $10{,}751$-trajectory corpus (\textbf{D1}; Table~\ref{tab:main_results}), our model aligns closely with ground truth (stepwise accuracy $89.44\%$, F1 $90.40\%$) while maintaining near-perfect legality ($99.99\%$), underscoring the value of explicit graph constraints layered on Transformer decoding.
In \textbf{D2}, lattice plans compile to phase-conditioned grasp--move--release primitives on an \emph{xArm Lite\,6}, achieving high success in central workspace ranges (Reach up to $97.5\%$, Pick up to $92.5\%$; Table~\ref{tab:rr_success}); rare failures are predominantly mechanical and are mitigated by twin re-grounding.
In \textbf{D3}, end-to-end execution over $60$ language-specified tasks attains $86.7\%$ overall success with robust handling of slips, occlusions, and incidental obstacles (Table~\ref{tab:robot_results}); the failure distribution (Table~\ref{tab:failures}) is dominated by state-sync and heavy occlusion.

\noindent\textbf{Why structure helps.}
The 3-grid representation encodes \emph{where/what/when}, and legality masks enforce unit-adjacent moves within workspace bounds, ensuring that decoded steps are physically feasible by construction.
This structural prior carries through to hardware: in \textbf{D2} and \textbf{D3}, legality prevents multi-axis jumps, while the digital twin converts symbolic nodes into calibrated targets and supports local pose re-grounding, obviating costly global re-plans.

\noindent\textbf{Observed error modes and their loci.}
Across \textbf{D1}, residual errors concentrate on (E1) tail truncation for long horizons and (E2) adjacent-step swaps; (E3) boundary nudges arise from score ties between neighboring valid nodes.
In \textbf{D2}, failures are mostly mechanical (e.g., transient gripper slip on glossy plastic) and are quickly recovered via target updates (e.g., after a $15^\circ$ rotation).
In \textbf{D3}, the principal failure modes are state-sync ``no state'' (8.3\%), occlusion/tight clustering (1.7\%), small nested items blocking grasps (1.7\%), and perception mis-ID (1.7\%)---see Table~\ref{tab:failures}.

\noindent\textbf{Implications.}
The combination of constraint-masked decoding and twin-backed execution yields \emph{reliable symbolic$\to$motor grounding} (D2) and \emph{robustness in clutter via local re-grounding} (D3), delivering $\sim 87\%$ task-level success without global re-planning.
These results suggest that graph-constrained Transformers can serve as dependable high-level planners whose outputs remain executable after sim$\to$real transfer.

\noindent\textbf{Limitations and remedies (by design).}
For \textbf{D1}, horizon-aware decoding (coverage penalties, scheduled sampling, curriculum for long paths) can reduce truncation and swaps, and explicit boundary handling can further trim nudges.
For \textbf{D2}, mechanical slips motivate improved grasp-force control and material-aware grasp selection.
For \textbf{D3}, the observed failures point to (i) multi-view/active perception to reduce ``no state'' events, (ii) target disambiguation and confidence-gated execution to curb mis-ID, and (iii) grasp reordering with nested-occupancy checks to address occlusion and container clutter.

\noindent\textbf{Outlook.}
A tighter loop between perception and decoding (frame-majority voting, uncertainty-aware masks), plus adaptive grasping and local clearing primitives, should raise success rates and cut retries in \textbf{D3} while preserving the structural guarantees validated in \textbf{D1} and the motor reliability demonstrated in \textbf{D2}.
Overall, \emph{PathFormer} illustrates that integrating graph structure with sequence models, and coupling them to a digital twin, provides a scalable route to resilient sim-to-real robot behavior.

\section{Conclusion}
\label{sec:conclusion}

We introduced a \emph{Path-based Transformer} that unifies spatial motion, task structure, and action order through a 3-grid representation (\emph{where/what/when}) and constraint-masked decoding. Embedding graph structure directly into sequence prediction yields trajectories that are accurate, interpretable, and physically feasible. On the $10{,}751$-trajectory corpus (D1), the model attains strong sequence fidelity (stepwise accuracy $89.44\%$, F1 $90.40\%$) with near-perfect legality ($99.99\%$). When compiled to motor primitives on hardware (D2), central-workspace success peaks at $97.5\%$ for \emph{Reach} and $92.5\%$ for \emph{Pick}; in real, cluttered scenes with language goals (D3), end-to-end success reaches $86.7\%$ while absorbing slips, occlusions, and incidental obstacles through local re-grounding rather than global re-plans.
By bridging sequence models and graph-based planning, the approach provides symbolic$\to$motor reliability: legality masks prevent multi-axis jumps, and the digital twin converts lattice nodes into calibrated targets and supports rapid pose updates, enabling robust sim$\to$real behavior while preserving safety envelopes.

We will extend the method to longer horizons and multi-task settings using horizon-aware decoding—with coverage penalties, scheduled sampling, and explicit boundary handling—and tighten the perception–decoding loop via uncertainty-aware masks, multi-view or active perception, and confidence-gated execution to reduce state-sync failures. We also plan to incorporate grasp reordering with nested-occupancy checks to handle container clutter and to couple PathFormer with high-level symbolic or language-conditioned planners alongside MPC-style refinement, thereby marrying task-level reasoning with low-level safety.
Structurally constrained Transformers, coupled to a digital twin, offer a scalable path to reliable, constraint-aware trajectory generation and execution for general-purpose robotic manipulation.

{
    \small
    \bibliographystyle{IEEEtran}
    \bibliography{references}
}

\end{document}